\documentclass{article}
\usepackage{spconf,amsmath,graphicx}

\AtBeginDocument{%
  }

\usepackage{amsmath, graphicx}
\usepackage{xcolor}
\newcommand{\revision}[1]{\textcolor{black}{#1}}

\usepackage{comment}
\usepackage{amsmath}
\DeclareMathOperator*{\argmaxA}{arg\,max} 

\usepackage{multicol}
\usepackage{booktabs,multirow}       

\begin{document}

\title{GPT-Calls: Enhancing Call Segmentation and Tagging by Generating Synthetic Conversations via Large Language Models}





\name{Itzik Malkiel$^{1}$\thanks{$^{*}$ Denotes equal contribution.}$^{*}$~~Uri Alon$^{1*}$~~Yakir Yehuda$^{1,2}$~~Shahar Keren$^{1}$~~Oren Barkan$^{1,3}$~~Royi Ronen$^{1}$~~Noam Koenigstein$^{1,4}$
}
\address{$^1$Microsoft, $^2$Technion, $^3$The Open University, $^4$Tel-Aviv University}

\maketitle

\begin{abstract}
Transcriptions of phone calls are of significant value across diverse fields, such as sales, customer service, healthcare, and law enforcement. Nevertheless, the analysis of these recorded conversations can be an arduous and time-intensive process, especially when dealing with extended or multifaceted dialogues. In this work, we propose a novel method, GPT-distilled Calls Segmentation and Tagging (GPT-Calls), for efficient and accurate call segmentation and topic extraction. GPT-Calls is composed of offline and online phases. The offline phase is applied once to a given list of topics and involves generating a distribution of synthetic sentences for each topic using a GPT model and extracting anchor vectors. The online phase is applied to every call separately and scores the similarity between the transcripted conversation and the topic anchors found in the offline phase. Then, time domain analysis is applied to the similarity scores to group utterances into segments and tag them with topics. The proposed paradigm provides an accurate and efficient method for call segmentation and topic extraction that does not require labeled data, thus making it a versatile approach applicable to various domains. Our algorithm operates in production under Dynamics 365 Sales Conversation Intelligence, and our research is based on real sales conversations gathered from various Dynamics 365 Sales tenants.
\end{abstract}

\section{Introduction}
\label{sec:intro}

\begin{figure*}[t]
\includegraphics[width=0.99\linewidth]{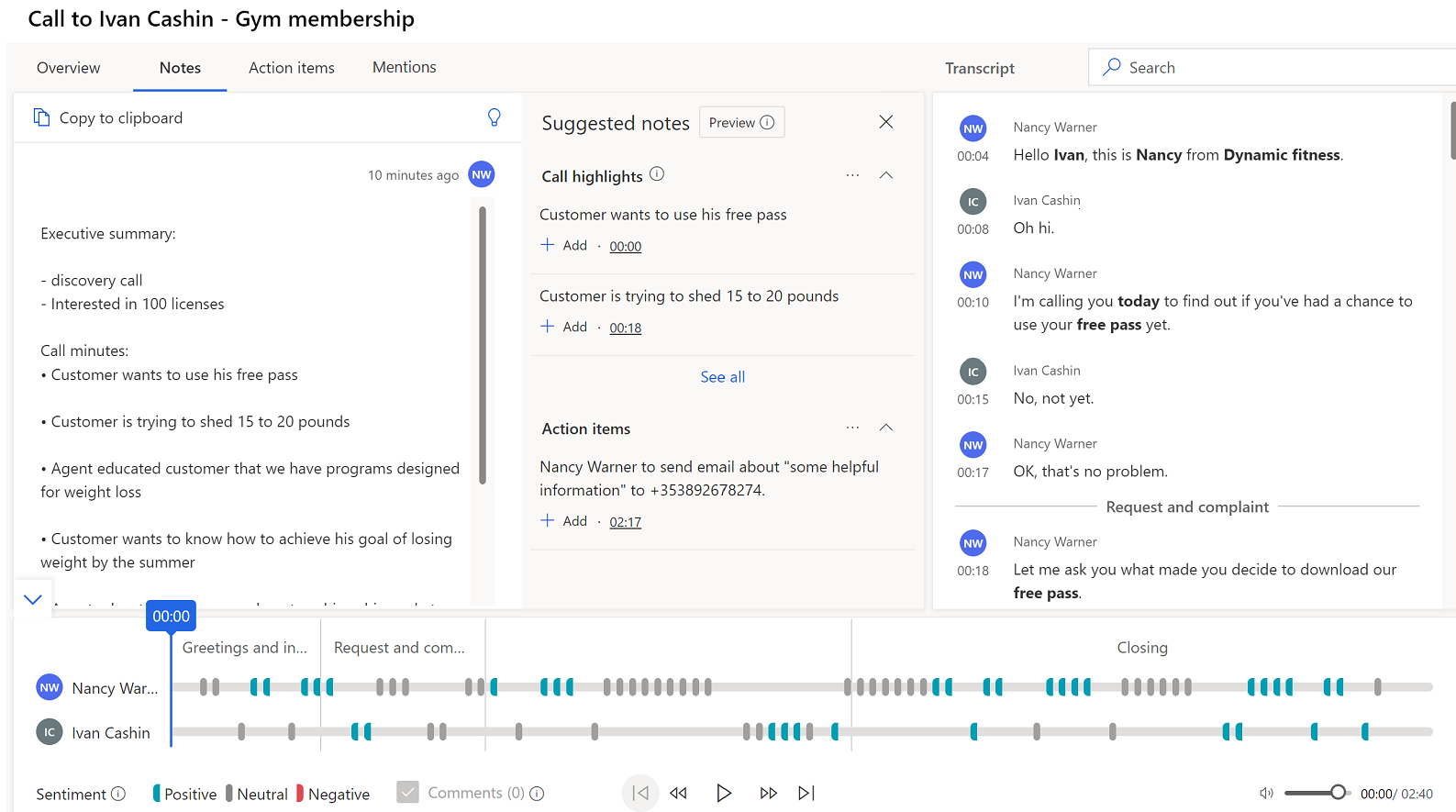}
\centering
\caption{The Calls Summary dashboard. Executive summary (left), suggested notes (middle), and the transcripted conversation (right) are presented. The predicted segmentation and their associated topics are showcased at the bottom, integrated with a corresponding timeline bar representing the duration of the recorded call.
}
\label{fig:sales365}
\end{figure*}

In today's highly competitive market, sales agents play a critical role in driving sales and maintaining a strong customer base. One of the primary ways they interact with customers is through phone calls. These phone calls are often recorded and transcribed for quality assurance, training and coaching other sellers, curating insights, and more. Therefore, sales departments often maintain an extensive database consisting of millions of call transcriptions, which serves as an important source of information for numerous tasks. 
However, analyzing recorded calls can be a challenging and time-consuming task, particularly when the conversations are lengthy or cover multiple topics. 

A common approach is to segment calls by post-processing the recorded text and assigning a topic to each segment. This simplifies the process for sellers to locate and extract important information from previous calls. Call segmentation involves breaking down conversations into smaller sections based on specific topics or themes discussed. This segmentation and tag capability greatly facilitates the day-to-day tasks of sellers and their managers. It enables sales agents and managers to easily track and analyze past calls, categorize them based on conversation topics, navigate to relevant parts of a call to extract crucial information, improve search engines that work with transcribed calls, and more.

Moreover, automated segmentation and tagging of recorded calls can help businesses optimize their sales strategies and processes across several dimensions:
\begin{itemize}

\item Providing personalized coaching to sellers. Managers can listen to specific parts of the call to provide feedback and guidance, rather than listening to the entire call.

\item Providing insights into customer needs, preferences, and pain points. Segmented and tagged information can be used to tailor sales strategies and improve customer experience.

\item Monitoring compliance with legal and ethical guidelines in sales conversations. Ensuring that sellers are adhering to company policies and regulations.

\item Generating reports and dashboards that provide insights into sales performance and customer behavior. This information can be used to make data-driven decisions and improve overall business outcomes.

\item Identifying key moments in the call that may be related to successful outcomes. This information can be used to train sales agents and improve their performance.
\end{itemize} 
Ultimately, all of the above lead to increased customer satisfaction, loyalty, sales, and revenue generation. 
Unfortunately, despite many advances in the field, current techniques for call segmentation and tagging have limitations that hinder their impact and penetration on the day-to-day work of sellers and their managers. Specifically, they often produce sub-optimal accuracy, struggle to represent various topics, and require significant labeled data and domain expertise to produce accurate ground truth segmentation. As a result, businesses have been slow to adopt post-processing segmentation in recorded calls.

\begin{figure*}[t]
\includegraphics[width=0.98\linewidth]{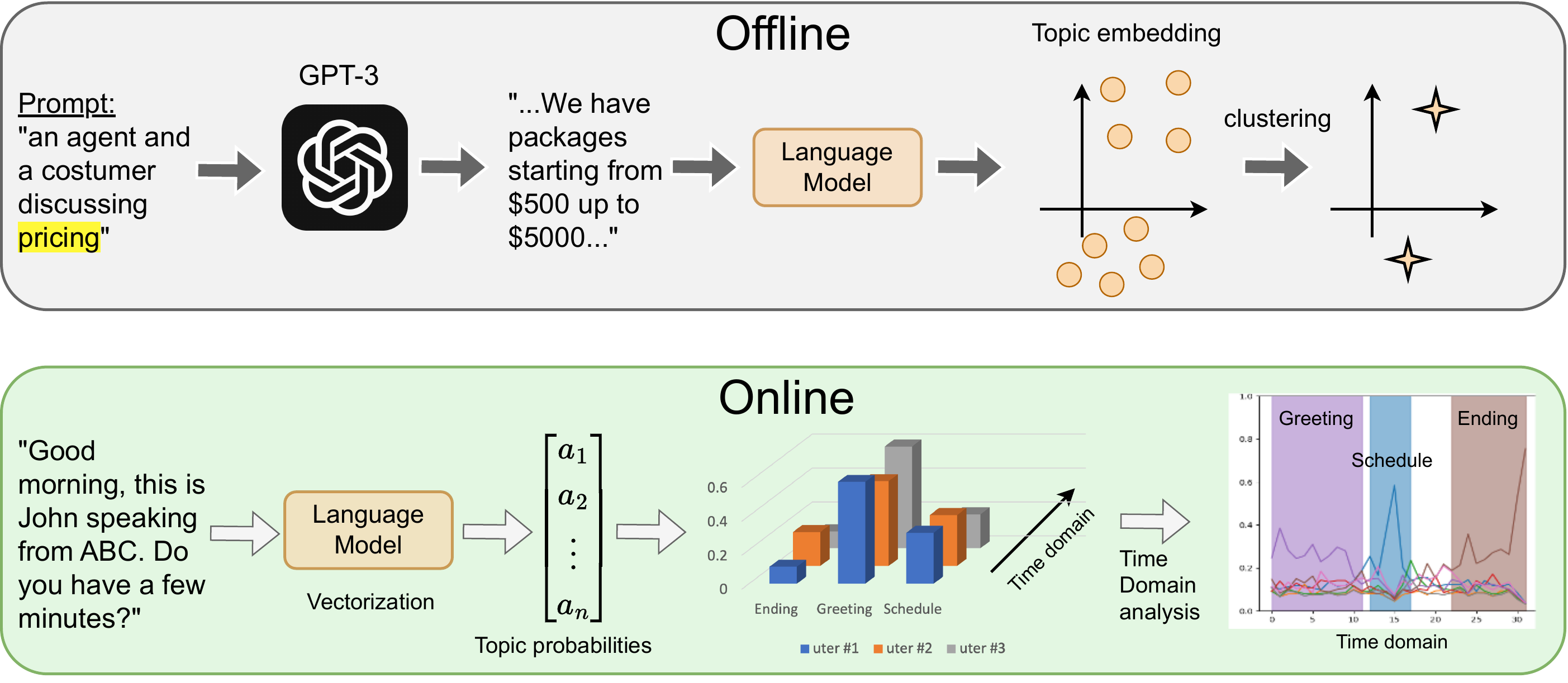}
\centering
\caption{An illustration of the offline and online phases of GPT-Calls.}
\label{fig:offline_online}
\end{figure*}

In this work, we propose a method that overcomes these limitations and provides accurate call segmentation and topic extraction without the need for labeled data. 
Based on a GPT model, our method generates synthetic data that mimics the natural language used in recorded calls, while focusing on a pre-defined set of 
topics. The synthetic data is then used by a smaller Transformer network to accurately segment calls and tag them with topics.

Our method, which we refer to as 
GPT-Calls, is designed to efficiently identify the topics discussed during a phone conversation by analyzing the similarity between the conversation utterances and predefined topic anchors. The algorithm consists of offline and online phases. In the offline phase, GPT-Calls generates a distribution of synthetic sentences for each topic using a GPT model~\cite{brown2020language, ouyang2022training}. Then, it embeds and clusters the sentences separately to extract anchor vectors. In the online phase, GPT-Calls processes a conversation transcript by embedding each utterance using a language model and scoring it for similarity with the topic anchors. Then, it applies time domain analysis over the scores of consecutive utterances and groups them into segments that are distinctly identified with a particular topic. The remaining segments that do not correlate with any of the topics are considered ``background''.

The GPT-Calls method offers an effective and precise solution for call segmentation and topic extraction that can cater to any topic without the need for labeled data. In this way, organizations can select a list of customized topics that are relevant to their particular business needs. Additionally, the algorithm is generic and can be implemented in diverse domains, such as customer support, sales conversations, surveys, and more. Finally, the online phase of the system can be applied in near-real-time scenarios, enabling the segmentation and tagging of an entire call within two-three seconds using standard hardware.
The GPT-Calls was recently adopted in Dynamics 365 Sales, and its predicted segmentations are apparent in the lower part of the Calls Summary dashboard (see Figure~\ref{fig:sales365}).

Our contributions are as follows: (1) we introduce the GPT-Calls scheme, a general method designed for the analysis of recorded calls in any domain. (2) we evaluate and report the performance of GPT-Calls on a diverse dataset of calls from multiple domains and tenants. (3) we compare our proposed method against other state-of-the-art alternatives. Our results demonstrate that the GPT-Calls method outperforms other approaches by a sizeable margin, and across all datasets.

\section{related work}
In recent years, there has been a growing interest in text segmentation, which involves dividing text passages into coherent segments based on their content. While traditional text segmentation methods have relied on features such as punctuation, paragraph breaks, or rule-based approaches, these methods may not always capture the underlying semantic structure of the text.

To address this challenge, several recent studies have proposed using semantic word embeddings to identify segments based on the coherence of the relationships between words. One such study proposes a method for text segmentation based on semantic word embeddings~\cite{alemi2015text}. In this work, the authors use a pretrained word embedding model to generate embeddings for each word in the text and then apply a greedy algorithm to group the words into segments. The authors demonstrate that their approach outperforms traditional methods in terms of segmentation accuracy and that the identified segments correspond well to the topics discussed in the text. In the experiments section, we compare our method with the previous method used in Dynamics 365 Sales for call segmentation, which was based on the same approach of \cite{alemi2015text}.

A different direction has been to employ text summarization methods for topic tagging. This approach involves assigning one or more tags to a given text to represent its key topics or themes. One such approach, known as extractive summarization, involves extracting key sentences from the text and using them to tag the text with pre-defined topics. Another approach, known as abstractive summarization, employs neural-based summarization models, such as sequence-to-sequence models or transformer models, to generate concise summaries that can be used to predict relevant tags. The latter approach can utilize the PEGASUS summarization model \cite{zhang2019pegasus}, which generates summaries that can ease the process of predicting tags (compared to predicting tags for the original text). The previous method used in Dynamics 365 sales built upon the PEGASUS model to summarize the segment and infer a relevant topic. More details and evaluations for this method can be found in Sec.\ref{exp}.

TextTiling~\cite{hearst1997text} is a prominent text segmentation algorithm that efficiently divides long text into coherent topical sections. It utilizes local lexical cohesion and focuses on identifying abrupt shifts in the thematic structure of a text, which are indicative of topic boundaries. By employing statistical techniques such as tokenization, similarity measurement, and smoothing, TextTiling extracts informative features from the text and clusters them into distinct sections, allowing for better understanding and organization of large text corpora. This versatile segmentation algorithm has been widely applied in various natural language processing tasks, including information retrieval, summarization, and text classification, making it a valuable tool for researchers in the field of computational linguistics.

Topic segmentation of meetings has gained significant attention as a means to automatically partition meeting transcripts into coherent segments, facilitating a better understanding of the discourse structure. In \cite{solbiati2021unsupervised}, the authors propose an unsupervised method for meetings segmentation by BERT embeddings (BERT-TSeg). Their method builds upon BERT to generate contextualized word representations for each utterance in the transcript. By leveraging these embeddings, BERT-TSeg computes the similarity between adjacent segments and constructs a similarity matrix. Subsequently, a hierarchical clustering algorithm is applied to this matrix to identify topic boundaries. In this study, we conduct a comprehensive evaluation and comparison between our proposed method and BERT-TSeg to assess the advancements and efficacy of our approach in meeting topic segmentation.

\section{Method}
The GPT-Calls algorithm consists of two distinct phases: an offline phase and an online phase. 
The offline phase uses a GPT model to generate synthetic data. This phase is applied once for a user-defined list of topics. After the offline phase, the algorithm invokes an online phase that reinforces representations extracted during the offline phase to accurately predict segmentation and topics for a given call. 

\subsection{Offline Phase}
In the offline phase, the algorithm utilizes a GPT model to generate synthetic data, leveraging a given list of desired topics chosen by the user. This phase is executed only once for a specific set of desired topics. Subsequently, in the online phase, the algorithm uses the representations extracted during the offline phase to predict segmentation and topics for individual calls.

\begin{figure}[t]
\includegraphics[width=0.95\linewidth]{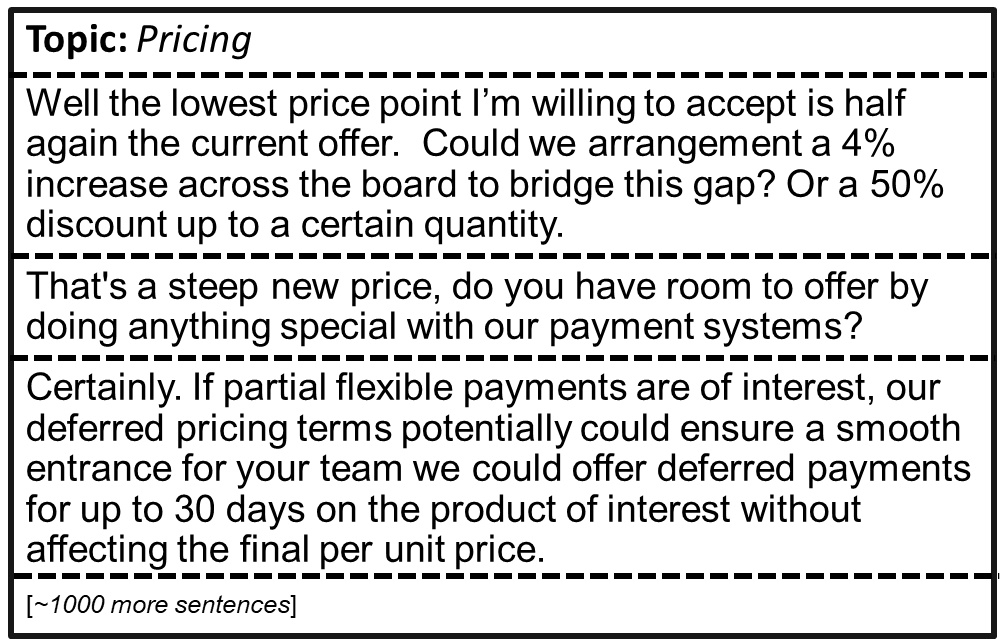}
\centering
\caption{Representative synthetic sentences generated by GPT-3 for the topic pricing.}
\label{fig:pricing_sentences}
\end{figure}

Specifically, for each topic, we build a prompt\footnote{The prompts contain one example of a few sentences and a topic which are followed by a request for the model to generate another example for the query topic} and use GPT-3 to generate thousands of sentences that are semantically correlated with the topic. 

A representative sample of sentences generated for the ``pricing'' topic is shown in Figure~\ref{fig:pricing_sentences}. An example of our prompts can be found in the Appendix Section~\ref{sup:prompt}.

Sentences produced for each subject are embedded via a pretrained Sentence-BERT (SBERT) model \cite{reimers2019sentence}. This model, an adjustment of the standard BERT model \cite{devlin2018bert}, is devised to encode entire sentences into a 384-dimensional embedding vector, enabling the use of the cosine similarity metric to measure semantic similarity.

Then, the DBSCAN \cite{ester1996density} algorithm is applied on the sentence embeddings of each topic, in order to extract a set of multiple ``anchors'' representing the distribution of the topic. DBSCAN is a density-based clustering algorithm that groups data into clusters based on the density of samples. High-density regions are grouped into a cluster, and samples in low-density areas are marked as outliers. For each topic, DBSCAN is applied to retrieve a set of clusters. The center of each cluster is extracted and used as an anchor. 

These anchors will be used during the online phase to infer the topic probabilities for each utterance in the call. The pipeline of the offline phase is illustrated in Figure~\ref{fig:offline_online} (upper part).

Overall, the offline phase involves defining topics, generating synthetic sentences, embedding them using SBERT, clustering the embeddings using DBSCAN, and extracting anchors. The anchors will be used later during the online phase to infer the probabilities of the topics.

\subsection{Online Phase}
In the online phase, GPT-Calls operates on the transcriptions of the recorded conversations and predicts the topic probabilities for each utterance in a given conversation. An utterance is an atomic unit of speech, which is mostly converted to a single sentence or a sub-sentence by the transcription model. 

\revision{The method employs the Azure Cognitive Service transcription model\footnote{https://azure.microsoft.com/en-us/products/cognitive-services/speech-to-text/} and embeds the resulting transcripted utterances through SBERT.}
GPT-Calls then iterates over the embedding of each transcripted utterance, scoring its similarity with all anchors of the pre-defined topics. For each topic, an utterance-topic score is defined by the maximal cosine similarity between the transcripted utterance embedding and the anchors associated with the topic (which are also vectors in the same latent space, as described in the offline phase). The utterance-topic scores for each topic are transformed into probabilities using the Softmax function. 
By performing this process, one obtains a sequence of vectors, where each vector represents the probability that the corresponding utterance relates to each of the topics.

To improve the accuracy of the topic probabilities, GPT-Calls applies a time-domain analysis to the above sequence, treating it as a multivariant time series. It identifies the peak points in each dimension of the time series, referred to as ``heat sources''. GPT-Calls then applies a heat diffusion strategy to the neighboring samples surrounding each heat source. For every sample in the sequence and across each dimension, GPT-Calls calculates the distance to the nearest right and left heat sources, strengthens the probability of the current sample's topic in proportion to the value of the closest right and left heat sources, and decays the probability by the distance. In other words, the probabilities of samples that are close to other samples that highly correlate with a specific topic will be slightly promoted toward the same topic. 

This approach intends to counteract the presence of noisy data samples which often manifest in particular topics such as identification.

After the heat diffusion process, a Softmax function is applied again to each utterance to assure its scores across the topics are valid probabilities that sum to 1. GPT-Calls then applies a sliding window technique to tag windows of consecutive utterances with topics.
Different window widths are utilized for each topic, which are hyperparameters determined individually for each topic using a validation set. The cumulative probability of a specific topic is computed for each window by averaging the relevant probabilities of all utterances within that window. If the cumulative probability exceeds a predefined threshold (configurable for each topic), the window is labeled with the corresponding topic. This process is repeated for all topics.

\begin{figure}[t]
\includegraphics[width=0.99\linewidth]{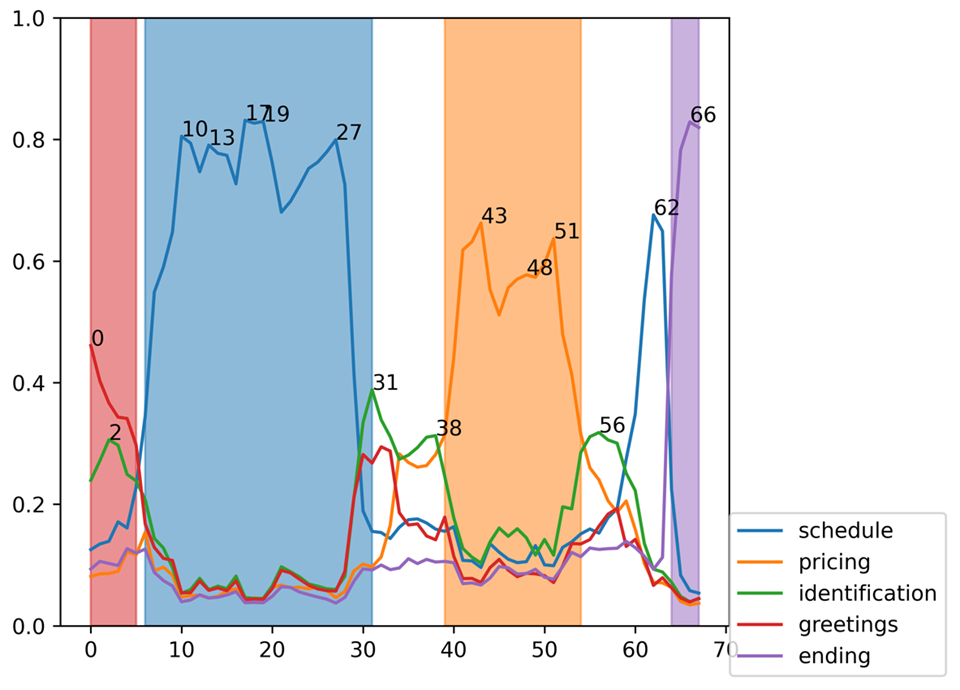}
\centering
\caption{Topic score (Y-axis) vs utterance index (X-axis) of a representative call with 67 utterances. All topic scores of all utterances are shown. Each topic score is presented by a solid line with a different color. The final retrieved segments are marked over the relevant utterances as a colored background by the corresponding topic color.}
\label{fig:raw_result}
\end{figure}

At the end of the above process, the sequence is associated with windows tagged with topics. GPT-Calls then iterates through the tagged windows and merges consecutive windows with the same topic. If a window or sub-window was tagged with more than one topic, the leading topic (with the highest score) is chosen, and the other windows are updated accordingly. The predicted segmentation and tagging are retrieved as the final output. A visualization of the predicted segmentation and tags along with the underlying probabilities of each utterance can be seen in Figure~\ref{fig:raw_result}. The pipeline of the online phase is illustrated in Figure~\ref{fig:offline_online} (bottom part).

\begin{table*}[t!]
\centering

\begin{tabular}{lcccccc}
\toprule
&\multicolumn{2}{c}{Sports} & \multicolumn{2}{c}{IT} & \multicolumn{2}{c}{Diverse}\\
\cmidrule(lr){2-3}  \cmidrule(lr){4-5}  \cmidrule(lr){6-7} 

&{\small Pk} & {\small WinDiff} & {\small Pk} & {\small WinDiff} & {\small Pk} & {\small WinDiff} \\
\midrule
TextTiling & 0.66 $\pm$ 0.07 & 0.89 $\pm$ 0.12  
 
& 0.65 $\pm$ 0.08 & 0.93 $\pm$ 0.12 & 0.66 $\pm$ 0.10 & 0.92 $\pm$ 0.12\\
BERT-TSeg & 0.36 $\pm$ 0.08  &0.36 $\pm$ 0.08  &

0.34 $\pm$ 0.10 & 0.35 $\pm$ 0.10 & 0.34 $\pm$ 0.10 & 0.35 $\pm$ 0.10\\
GSGST & 0.33 $\pm$ 0.09  &0.34 $\pm$ 0.08  &

0.31 $\pm$ 0.12 & 0.32 $\pm$ 0.11 & 0.32 $\pm$ 0.11 & {0.33 $\pm$ 0.11} \\
GPT-Calls & \textbf{0.29 $\pm$ 0.13}  & \textbf{0.31 $\pm$ 0.13}  &

\textbf{0.29 $\pm$ 0.10} 

& \textbf{0.30 $\pm$ 0.10}   
& \textbf{0.29 $\pm$ 0.14} 
&\textbf{ 0.31 $\pm$ 0.11} 
\\
\bottomrule
\end{tabular}
\caption{Pk and WinDiff scores for each model and dataset, reported as the mean values with standard deviations (Mean $\pm$ SD).}
\label{tab:tenants}
\end{table*}

\section{Experiments}
\label{exp}

In our evaluation, we compare GPT-Calls with baseline methods mentioned in the related work section, including TextTiling \cite{hearst1997text} and the unsupervised BERT-TSeg method \cite{solbiati2021unsupervised}. TextTiling employs a sliding window approach based on lexical cohesion to identify coherent segments within a document. BERT-TSeg, on the other hand, utilizes BERT embeddings and hierarchical clustering to identify topic boundaries in meeting transcripts. 

We also compare to the previous model used in Dynamics 365 Sales, which employed a technique that utilizes a \textbf{\underline{G}}reedy \textbf{\underline{S}}egmentation approach followed by a \textbf{\underline{G}}PT-based distilled \textbf{\underline{S}}ummary for \textbf{\underline{T}}agging (GSGST). GSGST first applies a segmentation procedure and then tags each segment with a relevant topic. The segmentation begins by embedding all the utterances in a given call using a pre-trained SBERT model. Then, the segments are inferred by employing the greedy method introduced in \cite{alemi2015text}\footnote{The implementation can be found here https://github.com/chschock/textsplit} over the utterances embedding. Given a call with $N$ embedded utterances $(w_1,...,w_N)$, a segment $V = (w_b,...,w_e)$ where $b,e$ are the beginning and ending indices\footnote{The minimal size of a segment is $3$} of $V$ ($0 \leq b + 1 <e \leq N$)
, and a split index $b<t<e$, the gain of splitting $V$ at position $t$ is defined as: 

\begin{equation*}
g_{b}^{e}(t) :=
\left\Vert
\sum_{i=b}^{t-1} w_i \right\Vert + \left\Vert
\sum_{i=t}^{e} w_i \right\Vert - \left\Vert
\sum_{i=b}^{e} w_i \right\Vert
\end{equation*}

The greedy approach calculates the index that maximizes this term 
\begin{align}
t^*:= \argmaxA_{t} g_{b}^{e}(t)
\end{align}
Then, if $g_{b}^{e}(t^*) > \tau $ where $\tau > 0$ is a pre-defined threshold, the segment $V$ is splited into two segments by around $t^*$. 

The method begins by initiating $b$ and $e$ as the first and last utterances of the conversation respectively. The process proceeds recursively to all segments and stops when there is no candidate split whose gain is above the pre-defined threshold, or the current number of segments has reached the maximum defined by the user. Finally, the last segmentation is retrieved as an output.

Given a predicted segmentation, two models are employed to predict a relevant topic, a zero-shot model introduced in \cite{yin2019benchmarking} and a summarization model based on PEGASUS \cite{zhang2019pegasus}. If the first model fails to predict a topic with high confidence, the second model is used.

The second model is based on the PEGASUS model, which was fine-tuned on sales calls where the labels were summaries generated by GPT \cite{asi2022end}.

The method assumes that each topic is associated with a small set of representative sentences, created by an expert (typically around 2-20 sentences). For example, for the topic ``pricing'' the corresponding set contains 9 sentences, two of them are: “the agent and customer discussed the price of the product”, “the customer asked for a better price”, and so on. These sentences were separately embedded by an SBERT model and were mean pooled to extract a single anchor representing the topic.

This model produces a single-sentence summary for every segment. The summarized segment is then embedded using SBERT and is compared to the single anchor of each topic. The predicted topic is the one that maximizes the cosine similarity with the summarized sentence embedding. Finally, post-processing is performed to filter out extremely short segments, merge adjacent segments with identical topics, and so on.

\begin{table*}[t] 
\centering
\label{tab:comparison}
\begin{tabular}{lccccc}
\hline
\textbf{} & \textbf{Identification} & \textbf{Pricing} & \textbf{Schedule} & \textbf{Greetings}  & \textbf{Closing} \\
\midrule
\textbf{GSGST} & 0.56/0.18 & 0.49/0.26 & 0.36/0.21 & 0.08/\textbf{0.04}  & \textbf{0.07}/0.05\\
\textbf{GPT-Calls} & \textbf{0.11/0.10}& \textbf{0.32/0.25 }&\textbf{ 0.20/0.15} & \textbf{0.07/0.04}  & \textbf{0.07/0.03}\\
\textbf{Improv.} & +80.4\%/+44.4\% & +34.6\%/+3.8\% & +44.4\%/+28.5\%  & +12.5\%/+0\%& +0\%/+40.0\%\\ \hline
\end{tabular}
\caption{ 
A comparison of the performance of the proposed method and the GSGST baseline, evaluated on the test set, for each topic separately.
Pk score/WinDiff are reported. Lower is better for both.}
\label{tab:results}
\end{table*}

\subsection{Metrics}
To evaluate the performance of the models, we calculate two metrics, Pk score~\cite{beeferman1999statistical} and WindowDiff~\cite{pevzner2002critique}. 

In order to measure the performance on each topic separately, the metrics are computed over the binary segmentation obtained by the predicted and ground truth segmentation of each of the topics. For example, when evaluating the topic ``pricing'', all the segments that are not associated with pricing were considered as background.
The Pk score is computed using a sliding window-based method, where the window size was set to half of the average true segment number. The metric determines whether the two ends of the window were in the same or different segments in the ground truth segmentation, and increased a counter if there was a mismatch. The final score is calculated by scaling the penalty between 0 and 1 and dividing the number of measurements.

To overcome the challenges of the Pk score (which penalizes false negatives more than false positives and does not take into account the number of boundaries), we also calculated the WindowDiff metric. Similar to the Pk score, the WindowDiff is also calculated using a sliding window. For each position of the window of size k, the number of boundaries in the ground truth segmentation is compared to the number of boundaries predicted by the model.

\subsection{Test data}
We employed a human expert with relevant domain expertise to annotate three test sets of real conversations taken from various Dynamics 365 sales tenants\footnote{with the tenant's consent and after all personally identifiable information was removed}. The human expert segmented each conversation and annotate each segment with one of the following topics: greetings, closing, pricing, identification, and scheduling. The annotations were used as the ground truth segmentation for evaluating the performance of the models. 

The first set, referred to as the ``\textit{\textbf{Sports}}'' set, contains $\sim 200$ sports-related conversation, where various sellers from sport companies discuss with their clients about subscription renewal, tickets ordering, seats changes and so on. The second set, named the ``\textit{\textbf{IT}}'' set, comprises $\sim 100$ recorded conversations of IT sellers reaching out to customers to propose software services, negotiate contracts, and address customer inquiries within the IT domain.  Lastly, the third set, labeled as the ``\textit{\textbf{Diverse}}'' set, consists of $\sim 200$ conversations of sale agents from various fields, including finance, technology, billing companies, medical marketing and more.

\subsection{Results}

Table \ref{tab:tenants} presents the segmentation performance of all baselines across the three datasets mentioned earlier, evaluated using the ground truth human annotations. In this evaluation, we focus solely on measuring the accuracy of the segmentation and do not consider the topics associated with the segments. As can be seen in the table, GPT-Calls outperforms all other alternatives by a sizeable margin. Specifically, compared to the second-best performing method, GSGST, we observe a relative improvement of $\sim12\%$, $\sim8\%$, in the Pk and WinDiff scores, respectively, for the Sports dataset. Furthermore, the proposed method shows even larger gains compared to the remaining baselines and across all three datasets.

\begin{table}[t]
\centering
\begin{tabular}{cccc}
\hline
\textbf{Model Name} & \textbf{Hit} & \textbf{Reasonable} & \textbf{Failure}\\
\midrule
GPT-Calls & 77.1\% & 16.2\% & 6.7\%\\
\hline
\end{tabular}
\caption{Human evaluation for the quality of segments. We measure the percentage of segments that (1) matches with the ground truth annotations (Hit), (2) mismatch with the ground truth but are considered as reasonable predictions (Reasonable), (3) mismatch with the ground truth and are not reasonable w.r.t. the underlying utterances in the segment (Failure).}
\label{tab:humaneval}
\end{table}

In Table \ref{tab:results}, we present our quantitative evaluations for all five topics: Identification, Pricing, Schedule, Greeting, and Closing. For these evaluations, we utilize the annotated test set and compare the performance of the GSGST model with our proposed model in terms of Pk score and WinDiff, separately for each topic. Both metrics aim to achieve lower values, indicating better performance.

The results demonstrate that our proposed model yields similar or better performance compared to the baseline model, across all topics. Specifically, for the Identification, Pricing, Schedule topics, we observe improvements ranging from 34.6\% to 80.4\% and 3.8\% to 44.4\% in in the Pk and WinDiff scores, respectively. The largest improvement was observed in the Identification topic, where our model achieved an 80.4\% improvement in the Pk score and a 44.4\% improvement in WinDiff.

\subsection{Human evaluations results}
We randomly selected 100 calls from the above three datasets, and evaluated the performance of the proposed model in an end-to-end manner, while focusing on the end-user experience. The model performance was evaluated using three criteria: (1) ``Hit'' was assigned when the predicted segment is well correlated with the ground truth segment. (2) ``Reasonable'' was designated when there was a discrepancy between the predicted segment and the ground truth, but the predicted segment and its topic are fairly associated with the underlying utterances. (3) ``Failure'' was determined when when the predicted segment did not match the ground truth, and the prediction did not align well with the underlying utterances.

The results, depicted in Table~\ref{tab:humaneval}, indicate that 93.3\% of the model predicted segmentations are considered fairly good (i.e. either ``Hit'' or ``Reasonable''), and only 6.7\% were detected as failures.

\section{Conclusion}
We propose a novel approach for call segmentation and tagging that builds upon distilling knowledge from the GPT-3 model and does not require labeled data. Our solution is generic and can be applied to  various domains. The proposed method is deployed in Dynamics 365 Sales Conversation Intelligence and was shown to significantly improve upon other alternatives.



\bibliographystyle{IEEEbib}
\bibliography{strings,refs}

\appendix
\section{Appendix}
\label{sup:prompt}

A single prompt was created for each topic and was used thousands times to generate a distribution of synthetic segments associated with the topic. The prompts contained one to four shots, enabling the model to focus on the task and generate high-quality synthetic segments. For instance, the pricing topic prompt is as follows:
``This is a prefix of a call between two people, where they greet and introduce each other: ``Thank you for calling Spencer and Bryce. This is Tracy. How can I help you? Hey Tracy, I'm Jeremy King from sales looking to reach Paul Lana. Uh, you know what? Give me your name again. Jeremy king. Calling regarding what Jeremy? I'm a salesperson working for'' 
Here is part of the middle of a different phone call between two different persons from different companies, where they are discussing the pricing of a product:''


\end{document}